\documentclass[letterpaper]{article} 
\usepackage[final, main]{neurips_2025} 
\usepackage{times}  
\usepackage{helvet}  
\usepackage{courier}  
\usepackage[hyphens]{url}  
\usepackage{graphicx} 
\usepackage{natbib}  
\usepackage{caption} 
\usepackage{algorithm}
\usepackage{algorithmic}
\usepackage{subcaption}
\usepackage{booktabs}
\usepackage{amsmath}
\usepackage[utf8]{inputenc} 
\usepackage[T1]{fontenc}    
\usepackage{hyperref}       
\usepackage{url}            
\usepackage{booktabs}       
\usepackage{amsfonts}       
\usepackage{nicefrac}       
\usepackage{microtype}      
\usepackage{xcolor}         
\usepackage{newfloat}
\usepackage{listings}
\DeclareCaptionStyle{ruled}{labelfont=normalfont,labelsep=colon,strut=off} 
\floatstyle{ruled}
\newfloat{listing}{tb}{lst}{}
\floatname{listing}{Listing}
\title{Alignment-Constrained Dynamic Pruning for LLMs: Identifying and Preserving Alignment-Critical Circuits}

\author{
\textbf{Dev Patel}\textsuperscript{1} \quad 
\textbf{Gabrielle Gervacio}\textsuperscript{1} \quad 
\textbf{Diekola Raimi}\textsuperscript{1} \quad 
\textbf{Kevin Zhu}\textsuperscript{1} \quad 
\textbf{Ryan Lagasse}\textsuperscript{1} \quad \\
\textbf{Gabriel Grand}\textsuperscript{2} \quad 
\textbf{Ashwinee Panda}\textsuperscript{3} \quad
\textbf{Maheep Chaudhary}\textsuperscript{4}\textsuperscript{\dag} \\[0.3em]
\textsuperscript{1}Algoverse \quad \textsuperscript{2}MIT \quad \textsuperscript{3}University of Maryland \quad \textsuperscript{4}Independent\\[0.3em]
\{writetodevp, maheepchaudhary.research\}@gmail.com \\[0.3em]
\textsuperscript{\dag}Project Lead \\
}

\begin{document}

\maketitle

\begin{abstract}
Large Language Models require substantial computational resources for inference, posing deployment challenges. 
While dynamic pruning offers superior efficiency over static methods through adaptive circuit selection, it exacerbates alignment degradation by retaining only input-dependent safety-critical circuit preservation across diverse inputs. 
As a result, addressing these heightened alignment vulnerabilities remains critical. 
We introduce Alignment-Aware Probe Pruning (AAPP), a dynamic structured pruning method that adaptively preserves alignment-relevant circuits during inference, building upon Probe Pruning. 
Experiments on LLaMA 2-7B, Qwen2.5-14B-Instruct, and Gemma-3-12B-IT show AAPP improves refusal rates by 50\% at matched compute, enabling efficient yet safety-preserving LLM deployment.
\end{abstract}


\section{Introduction}

\begin{figure}[h]   
    \centering
    \includegraphics[width=0.75\linewidth]{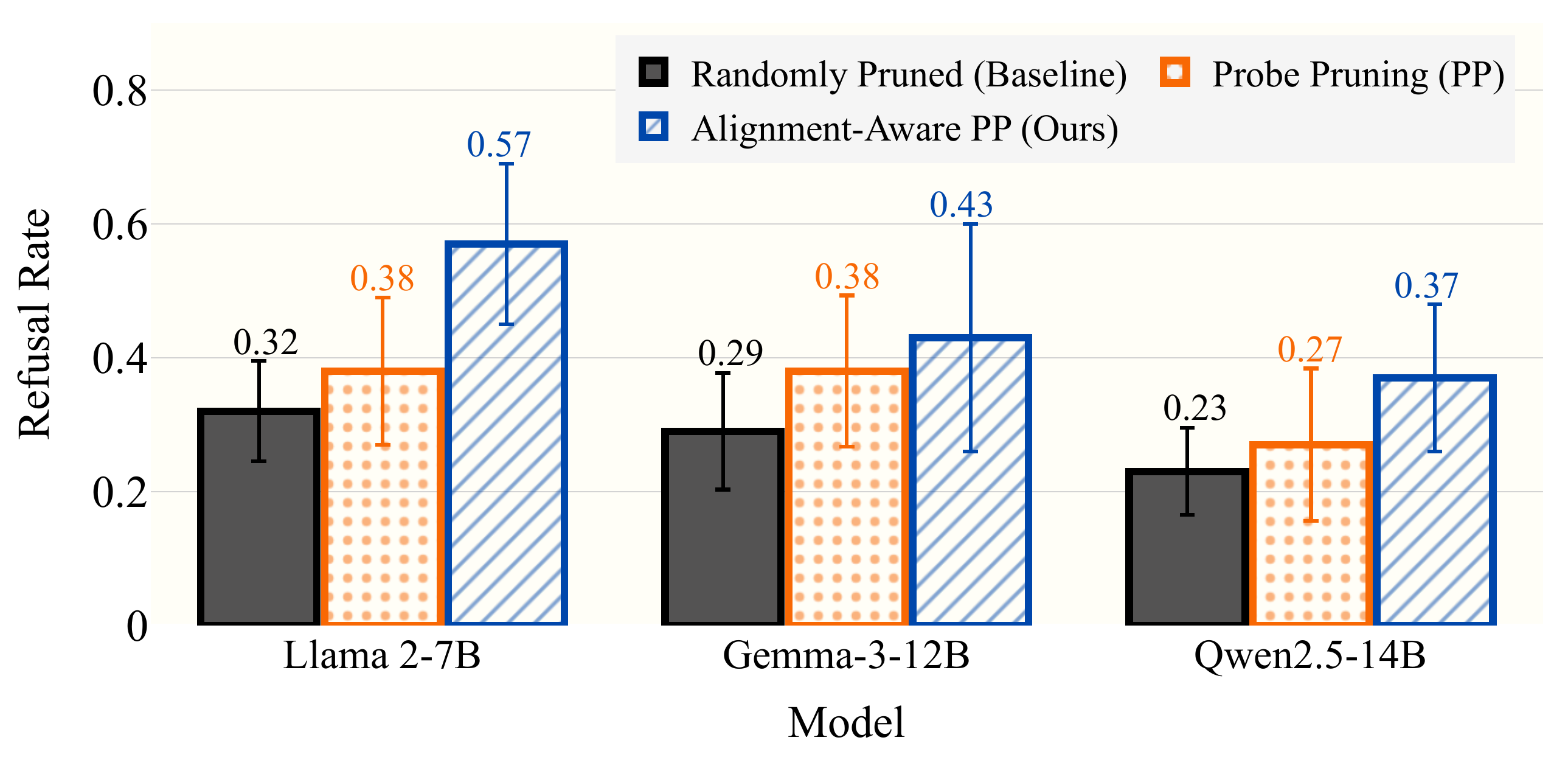}
    \caption{Refusal rates of LLaMA-2-7B, Qwen-2.5-14B, and Gemma-3-12B models on the WildJailbreak dataset \citep{jiang2024wildteaming} under pruning ratio $r=0.3$. We compare our Alignment-Aware Probe Pruning (AAPP) against two baselines: Probe Pruning (PP) \citep{le2025probepruning} and random pruning. Across all three models, AAPP consistently achieves higher refusal rates, demonstrating that preserving alignment-critical circuits upon the detection of adversarial prompts improves safety behavior under pruning.}
    \label{fig:refusal_rates}
\end{figure}

LLMs deliver impressive capabilities yet impose high computational costs, with inference costs scaling directly with model size \citep{kaplan2020scaling}. Pruning offers a promising route to reduce these costs \citep{han2016deep}, using different techniques, including static structured pruning \citep{ma2023llmpruner} as well as dynamic probe-guided pruning (PP) \citep{le2025probepruning} which improves the accuracy-efficiency frontier by pruning columns of the learnable linear transformation that maps intermediate hidden state to the output hidden state, referred to as an input channel. However, these methods risk pruning alignment-critical structures, potentially weakening safety guardrails and degrading behaviors such as refusal of harmful instructions. Recent analyses \citep{wei2024brittleness} show that removing as little as $3\%$ of parameters is enough to compromise safety. This brittleness motivates the development of Alignment-Aware Probe Pruning (AAPP)---a method that explicitly preserves alignment-critical circuits. 

AAPP uses the average activation value for each input channel. By comparing these scores obtained from benign and harmful prompts to the scores obtained from our probe pass, our method detects adversarial inputs and enforces hard exclusions on alignment-critical structures. This structured pruning approach yields an improved efficiency-alignment frontier: AAPP outperforms PP, having refusal rates up to 50\% greater for the same computational budget. These findings suggest constraint-satisfying pruning as a practical route to efficient yet safe LLMs.
Our key contributions are as follows:
\begin{itemize}
    \item We develop a pruning framework that preserves interpretable circuits
    \item We evaluate our framework on refusal rate, toxicity, accuracy, and computational cost (FLOPs)
\end{itemize}

\section{Related Work}

\subsection{Structured Pruning}
Structured pruning is a key approach for reducing the computational cost of LLMs. 
LLM-Pruner \citep{ma2023llmpruner} removes entire attention heads and MLP neurons via gradient-based importance, while Wanda \citep{sun2024wanda} prunes weights with small magnitude and activation values post-hoc, achieving high sparsity without retraining. Probe Pruning \citep{le2025probepruning} extends this line by using probed hidden states to guide batch-wise pruning, improving the accuracy-efficiency frontier. However, these methods risk pruning the preservation of alignment-critical structures.

\subsection{Alignment Preservation}
Several methods aim to preserve alignment by constraining intervening on the causal elements\citep{chaudhary2023towards, geiger2025causal} of the models responsible alignment during model modification. Safe LoRA \citep{hsu2024safelora} and SaLoRA \citep{li2025salora} constrain LoRA updates to remain within safety-aligned subspaces, while LoRI \citep{zhang2025lori} and LoTA \citep{panda2024lota} apply structural sparsity to reduce catastrophic forgetting. These works show that constraining fine-tuning helps preserve desirable behaviors in LLMs. NLSR \citep{yi2025nlsr} restores safety by transplanting safety-critical neurons from an aligned reference model. These approaches show that explicit parameter constraints and neuron transplantation can maintain refusal, honesty, and toxicity safeguards even under structural changes. Layer-level analyses further support targeted preservation: Shi et al.\ \citep{shi2024layers} showing that alignment changes concentrate in late-stage layers and that compression can focus on non-critical regions.

\section{Methods}

\begin{figure}[h]
  \centering
  \includegraphics[width=1\linewidth, height=0.37\linewidth]{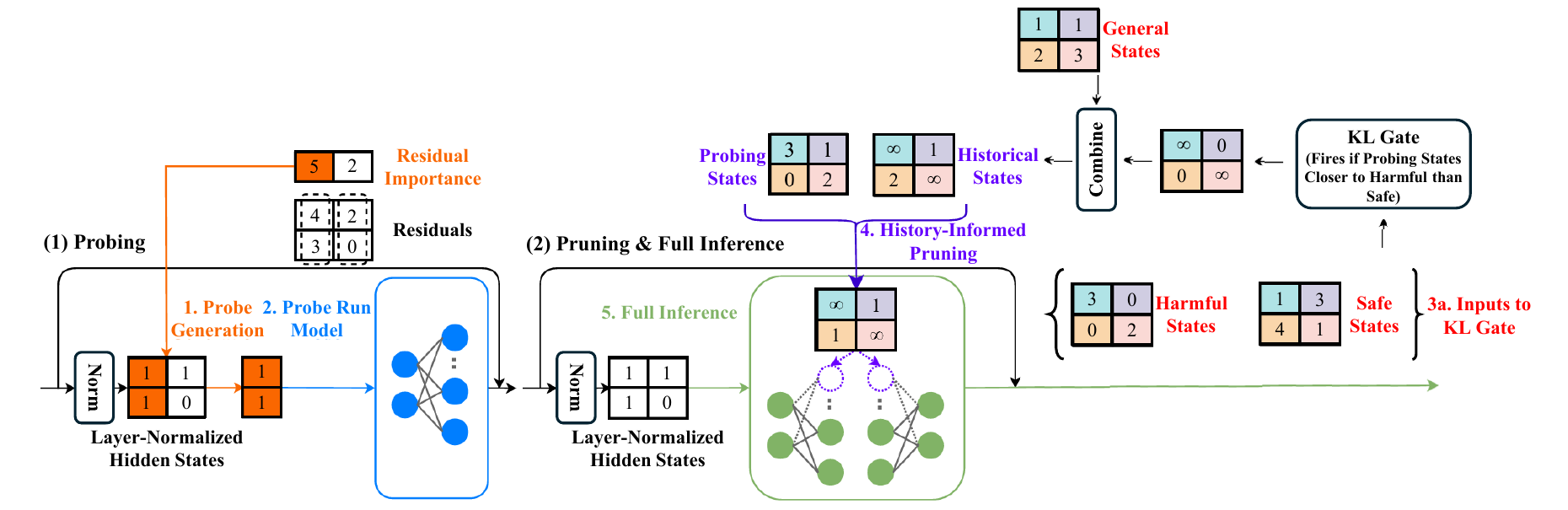} 
  \caption{Alignment-Aware Probe Pruning (PP) is executed in five stages: 
  (1) From the layer-normalized hidden states, pick tokens based on residual-importance and build a small probe.
  (2) Run the probe a few layers ahead to produce probing states
  (3a) A KL Gate compares them to historical states from safe and harmful prompts and fires when closer to harmful, ensuring the preservation of alignment-critical structures. If the gate does not fire, the probe states are just fused with the general historical states
  (4) Using the integrated states to calculate the pruning metric \citep{le2025probepruning}, prune low-score channels.
  (5) Perform full inference on the remaining weights.}
  \label{fig:pp-overview}
\end{figure}

As shown in Figure \ref{fig:pp-overview}, Alignment-Aware Probe Pruning consists of five stages, namely probe generation; probing, recording activations; comparison to our historical activation scores; history-informed pruning; and inference. 

\subsection{Activations and Scoring}

For each target with $C$ input channels, 
we create 3 tensors: general, benign, and
harmful using sets of prompts: (1) general prompts to maintain linguistic functionality from C4 dataset \citep{raffel2020exploring}; (2) benign prompts from wild adversarial dataset; and (3) harmful prompts from wild adversarial dataset. (\citep{jiang2024wildteaming}). Each set of scores stores the squared $\ell_2$ norm of channel activations compressed across the batch and sequence dimensions. We refer to this value as the ``channel's energy''.

For structured pruning, we adopt the PP\textsubscript{sp} importance metric from Probe Pruning~\citep{le2025probepruning}, which computes per-channel pruning scores using the $\ell_2$ norms of each input channel’s activations. Here, $W^{\text{final}}$ denotes the learnable linear transformation between hidden states, and $X^{\text{int}}$ the intermediate hidden state. A lower PP\textsubscript{sp} score, $I_k$, indicates less important channels.

\begin{equation}
I_k = \left\| \left\{ \, |W^{\text{final}}_{i,k}|^2 \cdot \| X^{\text{int}}_{:,:,k} \|_2^2 \, \right\}_{i=0}^{C_{\text{out}}} \right\|_2,
\end{equation}

Finally, we blend live scores with stored activation scores obtained from the set of general prompts. 

\subsection{Risk-aware gate and channel selection}

We keep $k = \lceil (1-r)C \rceil$ channels, reserving $k_{align} = \lfloor \text{align\_frac} \cdot C \rfloor$ channels for safety. 
Probing states; and historical states from benign and harmful prompts are normalized into distributions: `$p$'; and `$q_{\text{safe}}$, and $ q_{\text{jail}}$', respectively, using Equation \ref{eqn:one}.
\begin{equation} \label{eqn:one}
    KL_{\text{harm}} = \sum_c p_c \log \tfrac{p_c}{q^c_{\text{jail}}}, \quad
    KL_{\text{safe}} = \sum_c p_c \log \tfrac{p_c}{q^c_{\text{safe}}}.
\end{equation}
If $KL_{\text{harm}}-KL_{\text{safe}} \ge \tau_{\text{margin}}$, we preserve the top $k_{align}$ channels by $hist_{\text{jail}}$ as we wish to protect channels most active under harmful prompts because they include refusal circuitry. We then fill the remainder by descending score. Otherwise, we retain the top $k$ channels by score. Using these scores, binary masks are generated for pruning and then materialized to obtain real compute reductions.  

\section{Experimentation and Results}

We evaluate on HuggingFace implementations of Llama-2-7B-chat, Qwen2.5-14B-Instruct, and Gemma-3-12B-IT, using prompts from the WildJailbreak dataset (\citep{jiang2024wildteaming}) which were not used for the generation of historical states. Workloads contain prompts of avg.\ length 300 tokens with 120 tokens generated. Unless stated otherwise, we fix hyperparameters to align frac $=0.3$, refresh window $=20$, and batch size $=20$ for prompts.

We estimate inference FLOPs calculated using 2 FLOPs/MAC (\citep{hoffmann2022training}) taking into account the number of layers, attention heads, hidden size, intermediate size, and vocabulary size for the given model. We prune only in the input channels of attention $o_{\text{proj}}$ and MLP $down_{\text{proj}}$, excluding the first 6 and last 3 layers. Outputs are post-hoc labeled for refusal and toxicity. Metrics include throughput compute (FLOPs/token), refusal rate (trained classifier), classification accuracy and toxicity (Perspective API \citep{lees2022perspective}).  

Across the two methods (AAPP and PP), We first consider the model's ability to classify harmful and unharmful prompts and act accordingly. This is investigated across various compute budgets and prune ratios. Following this, we assess the safety of the model's responses for AAPP and PP using toxicity as the measure.

\subsection{Refusal Rates at Fixed Prune Ratio}

Figure \ref{fig:refusal_rates} presents refusal rates at prune ratio $r=0.3$. Across all three models, AAPP achieves higher refusal rates (implicit and explicit) than both Randomly Pruned and Probe Pruning (PP) baselines, preserving alignment behavior. On Llama-2-7B-chat, AAPP attains a refusal rate (0.57) 50\% and 78\% greater than PP (0.38) and Random Pruning (0.32), respectively. Similar improvements hold for Llama-2-7B-chat (37\% and 61\%) and Gemma-3-12B-IT (13\% and 48\%), confirming the robustness of our approach across architectures.

\subsection{Refusal Rates against Compute (FLOPs per Token)}

\begin{figure}[h]
    \centering
    \begin{subfigure}[b]{0.48\columnwidth}
        \centering
        \includegraphics[width=\textwidth]{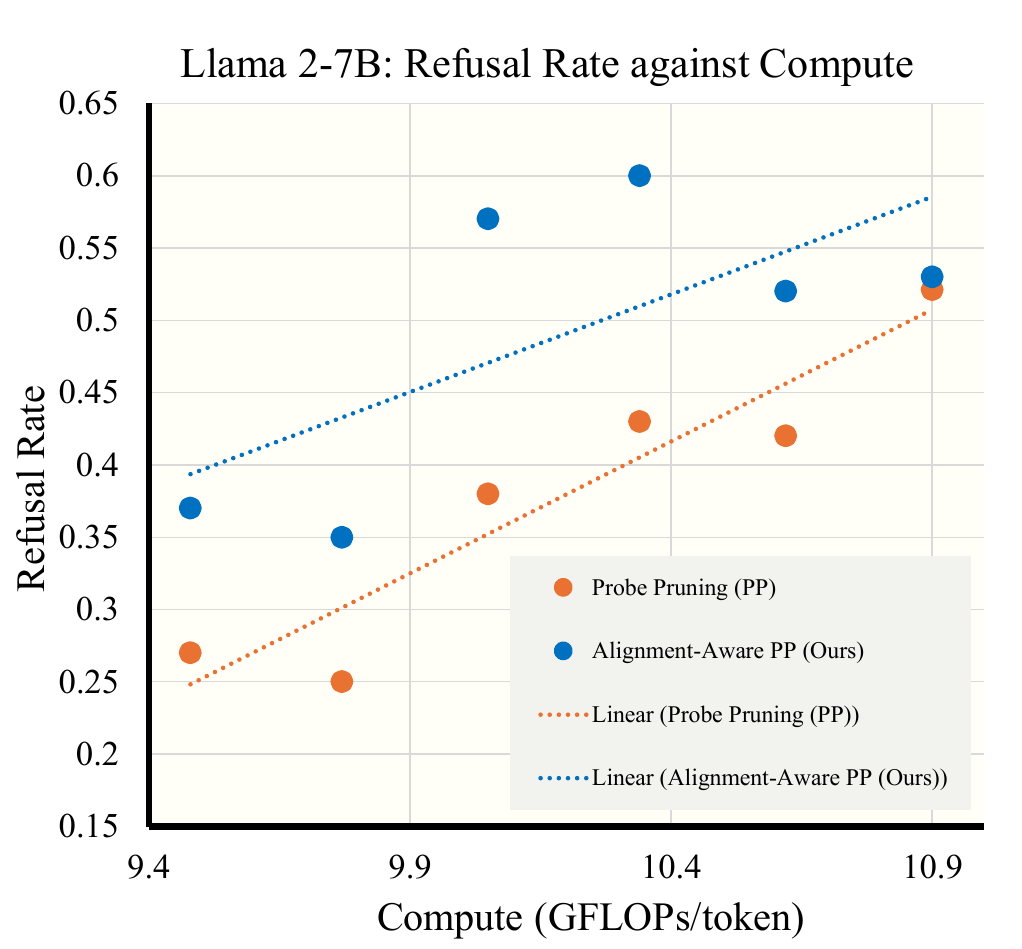}
        \caption{Llama-2-7B-chat. AAPP maintains substantially higher refusal rates at comparable compute budgets, achieving safer behavior with fewer FLOPs compared to standard PP.}
        \label{fig:refusal_vs_compute1}
    \end{subfigure}
    \hfill
    \begin{subfigure}[b]{0.48\columnwidth}
        \centering
        \includegraphics[width=\textwidth]{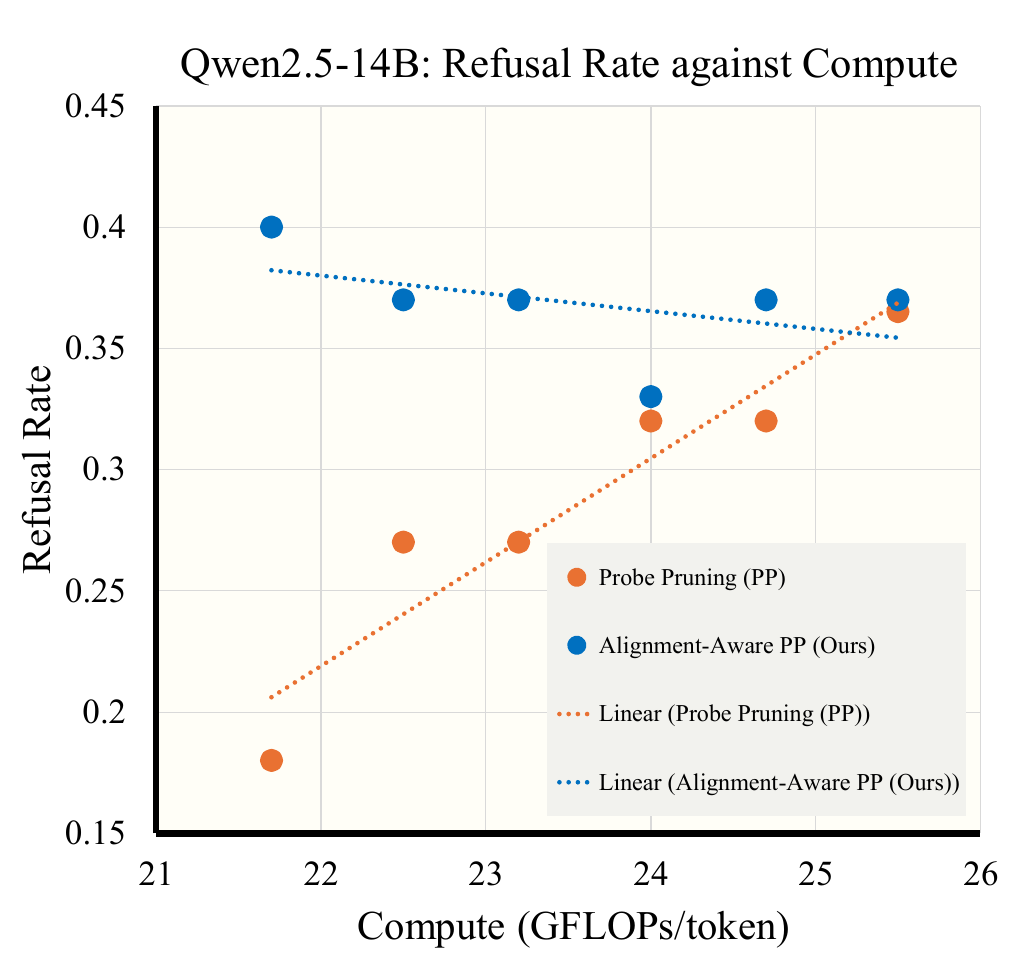}
        \caption{Qwen2.5-14B-Instruct. AAPP preserves refusal performance as compute decreases, improving the refusal-compute trade-off relative to PP across the efficiency spectrum.}
        \label{fig:refusal_vs_compute2}
    \end{subfigure}

    \caption{Refusal rate vs compute (GFLOPs/token) across models. AAPP consistently achieves higher refusal rates at lower compute costs than standard PP, demonstrating improved alignment–efficiency trade-offs.}
    \label{fig:refusal_vs_compute_combined}
\end{figure}

Extending the investigation, we vary compute budgets to look into the alignment-efficiency frontiers created using either method. Figure \ref{fig:refusal_vs_compute1} and \ref{fig:refusal_vs_compute2} illustrates alignment (refusal rate) as a function of computational efficiency (GFLOPs/token) for the Llama-2-7B-chat and Qwen2.5-14B-Instruct models, respectively, under Probe Pruning (PP) and Alignment-aware PP. Given the same computational budget, our method achieves a higher refusal rate, shifting the efficiency-alignment frontier upward. For example, on Llama-2-7B-chat (\ref{fig:refusal_vs_compute1}), to achieve a target refusal rate of 0.5, our method requires only 10.3\,GFLOPs/token, compared to a higher cost with PP. Qwen2.5-14B-Instruct exhibits the same pattern, demonstrating that AAPP maintains safety more efficiently across various compute levels. These results show that AAPP improves the alignment-efficiency trade-off, achieving safer behavior while reducing inference cost, and generalizing across diverse model families.

\subsection{Alignment Accuracy}
\begin{table*}[h]
\centering
\begin{tabular}{c c c ccc}
\toprule
\textbf{Model} & \textbf{Prune Ratio} & \textbf{Method} & \text{F1 ($\uparrow$)} & \text{Accuracy ($\uparrow$)} & \text{FAR ($\downarrow$)}\\
\midrule
& 0    & PP & 1.000   & 1.000   & 0.000 \\
     & & AAPP& 1.000   & 1.000   & 0.000 \\
Llama-2-7B-chat & 0.15 & PP & 0.725   & 0.702  & 0.290 \\
     & & AAPP & 0.834   & 0.808   & 0.201 \\
& 0.3  & PP & 0.645   & 0.624   & 0.313 \\
     & & AAPP & 0.760   & 0.741   & 0.254 \\
\midrule
& 0    & PP & 1.000   & 1.000   & 0.000 \\
     & & AAPP & 1.000   & 1.000   & 0.000 \\
Qwen2.5-14B-Instruct & 0.15 & PP & 0.876   & 0.891  & 0.058 \\
     & & AAPP &  0.880  & 0.916  & 0.05 \\
& 0.3  & PP & 0.730   & 0.820   & 0.169 \\
     & & AAPP & 0.786   & 0.858   & 0.092 \\
\bottomrule
\end{tabular}
\caption{Comparison of F1, Accuracy and FAR for PP and AAPP across prune ratios on Llama-2-7B-Chat and Qwen2.5-14B-Instruct: AAPP has a lower False Acceptance Rate with higher classification accuracy, behaving more similarly to the unpruned models.}
\label{tab:pp_aapp_prune}
\end{table*}

The accuracy of these refusals and the behavior of the model, more generally, is shown in Table \ref{tab:pp_aapp_prune}. It indicates that AAPP outperforms PP across prune ratios on Llama-2-7B-Chat and Qwen2.5-14B-Instruct. The results for the pruned models are compared to the unpruned model, which we consider to have a maximum for these metrics, as our pruned models cannot exceed the performance of the base model. We use F1 to balance recall and precision, accuracy and False Acceptance Rate to indicate how often the model does not refuse prompts. PP’s accuracy and F1 decline as pruning increases, dropping to 0.575 and 0.585 at a 0.3 ratio for Llama2-7B-Chat. In contrast, AAPP retains higher values, 0.741 accuracy and 0.760 F1, indicating stronger classification stability. Additionally, AAPP maintains a lower False Acceptance Rate (FAR) (e.g. 0.216 vs 0.353 at 0.3). Similar results can be seen for Qwen2.5-14B-Instruct. Overall, these results demonstrate AAPP’s ability to preserve safety and behavior near to the unpruned models at reduced compute.

\subsection{Toxicity against Prune Ratio}
\begin{figure}[h]
    \centering
    \begin{subfigure}[b]{0.48\columnwidth}
        \centering
        \includegraphics[width=\textwidth]{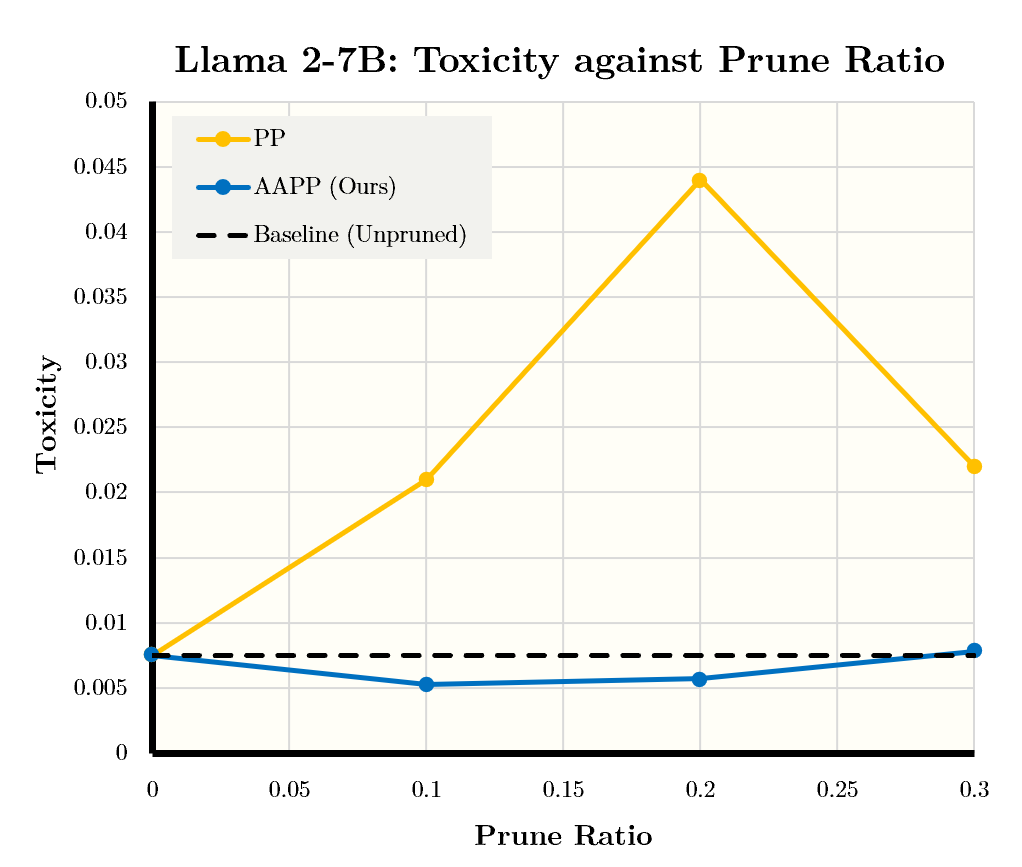}
        \caption{Llama-2-7B-chat. AAPP maintains toxicity levels closer to the unpruned baseline compared to PP, demonstrating better preservation of safety alignment under aggressive pruning.}
        \label{fig:tox1}
    \end{subfigure}
    \hfill
    \begin{subfigure}[b]{0.48\columnwidth}
        \centering
        \includegraphics[width=\textwidth]{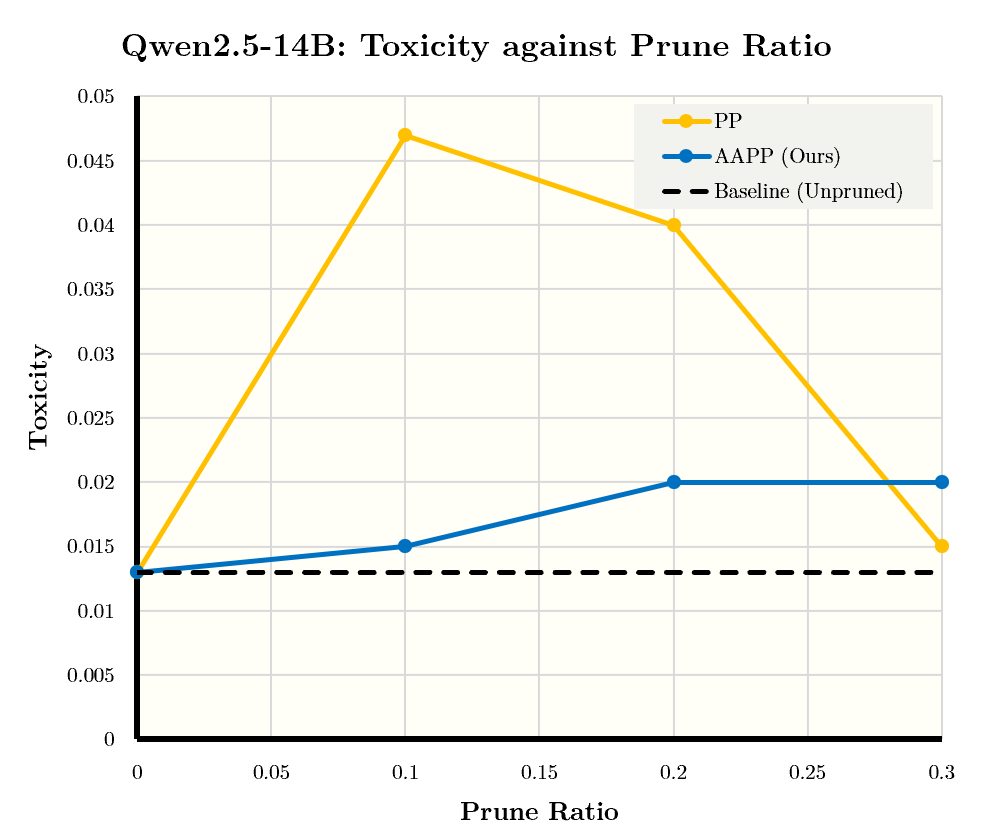}
        \caption{Qwen2.5-14B-Instruct. AAPP sustains lower toxicity scores closer to the unpruned model across pruning ratios, outperforming PP in safety preservation.}
        \label{fig:tox2}
    \end{subfigure}

    \caption{Toxicity vs prune ratio across models. AAPP consistently preserves lower toxicity and safer outputs under pruning, outperforming PP across both Llama-2-7B-chat and Qwen2.5-14B-Instruct.}
    \label{fig:tox_combined}
\end{figure}

Through toxicity, we can understand how safely the model responds. Figure \ref{fig:tox1} and \ref{fig:tox2} indicates that across both models, AAPP shows clear safety gains over PP. On Llama-2-7B-Chat, PP’s toxicity peaks at 0.044 at a 0.2 prune ratio, while AAPP stays nearly constant near 0.0075, matching the unpruned baseline. Similarly, on Qwen2.5-14B-Instruct, PP reaches 0.08, but AAPP remains below 0.02. This demonstrates that AAPP preserves alignment even under heavy pruning. Although toxicity scores decrease at high pruning ratios, this may reflect linguistic degradation rather than improved safety. Pruning can suppress expressive activations, yielding flatter, less coherent text that is rated as less toxic.

\section{Conclusion}
We propose a pruning method that preserves alignment while reducing inference cost. By integrating a risk-aware gate with probe-guided pruning, we prevent the removal of alignment-critical structures upon the input of an adversarial prompt and improves the efficiency-alignment frontier. Experiments on Llama-2-7B-chat, Qwen2.5-14B-Instruct, and Gemma-3-12B-IT show that AAPP sustains lower toxicity and greater classification accuracy at lower FLOP budgets, offering a practical route to safer and more efficient LLMs. 

Therefore, our method improves efficiency, scalability, and energy use without significantly compromising safety. However, there is a risk of missed unsafe inputs as the model is pruned, but we reduce the chance of this happening through conservative gating.

Limitations of our study include evaluation at mid-scale model sizes and approximate FLOP accounting. Future work will extend AAPP to larger models and investigate whether similar additions can be made to build upon probe pruning in other contexts.
\newpage
\bibliographystyle{plainnat}  
\bibliography{main}     

\end{document}